\documentclass[runningheads]{llncs}

\usepackage{booktabs}
\usepackage[T1]{fontenc}
\usepackage{graphicx}
\usepackage{amsmath,amssymb}
\usepackage{multirow}
\usepackage{array}
\usepackage{url}
\usepackage{xcolor}
\usepackage{hyperref}
\usepackage{algorithm}
\usepackage{algpseudocode}
\usepackage{fancyhdr}

\fancypagestyle{preprint}{
    \fancyhf{}

    \fancyhead[L]{%
        \normalfont\footnotesize
        Author-prepared version accepted at ICONIP 2026 (Springer CCIS).
    }

    \fancyfoot[C]{\normalfont\small\thepage}

}

\newcommand{\safeincludegraphics}[2][]{%
    \IfFileExists{#2}{%
        \includegraphics[#1]{#2}%
    }{%
        \fbox{%
            \parbox[c][3.1cm][c]{0.92\linewidth}{%
                \centering\small
                Missing figure:\\
                {\ttfamily\detokenize{#2}}%
            }%
        }%
    }%
}

\begin{document}

\title{Explainable AI for Olympic Medal Prediction:\\
A Two-Level Framework for Temporal Generalization and Sport-Specific Advantage}

\titlerunning{Explainable AI for Olympic Medal Prediction}

\author{
Yawen Lai\orcidID{0009-0006-3014-4859} \and
Jing Wei\thanks{Corresponding author.}\orcidID{0000-0002-6683-8522} \and
Yong Zhou
}

\authorrunning{Y. Lai, J. Wei, and Y. Zhou}

\institute{
School of Statistics, Chengdu University of Information Technology,
Chengdu 610103, China\\
\email{weijing@cuit.edu.cn}
}

\maketitle

% Apply the acceptance notice to every page, including the title page.
\pagestyle{preprint}
\thispagestyle{preprint}

% =============================================================================
% Abstract
% =============================================================================

\begin{abstract}

Olympic medal prediction requires not only accurate forecasting but also an
understanding of how national capacity is translated into sport-specific
success. We propose a temporally validated and explainable two-level framework
linking national medal capacity with sport-specific advantage. Module A models
national medal attainment and conditional medal accumulation, while Module B
extends the analysis to NOC--Games--Sport outcomes using sport-specific
participation, resource allocation, and competitive history. Models are
selected under expanding-window validation and evaluated on the withheld Paris
2024 Games. The national classifier achieves AP=0.9464, and the complete
two-stage system attains an end-to-end RMSE of 2.659 medals, reducing RMSE by
18.3\% relative to previous-Games persistence. At the sport level, AP improves
from 0.5172 to 0.7809 relative to previous-Games persistence. SHAP analysis shows
that delegation characteristics are particularly informative for medal
attainment, historical competitive capacity for medal accumulation, and
sport-specific allocation and history for sport-level success. Nonlinear
responses further reveal multi-level capacity thresholds in GDP, delegation
size, and sport-specific participation. $k$-means country profiling combined
with profile-centered SHAP analysis shows that higher- and
intermediate-capacity profiles exhibit distinct relative advantages across
sports. The framework provides an interpretable basis for benchmarking national
sporting capacity and understanding how resources are converted into
competitive advantage.


\end{abstract}

% =============================================================================
\section{Introduction}
\label{sec:introduction}
% =============================================================================

Accurate Olympic medal forecasting is a long-standing problem in sports
analytics, with practical relevance to competitive benchmarking, athlete
development, and strategic resource allocation~\cite{bernard2004,scelles2020}.
The task is challenging because medal outcomes are highly uneven across
countries, while participating nations differ substantially in economic
capacity, population, delegation size, sporting traditions, and accumulated
competitive experience. Moreover, Olympic performance is inherently
sport-dependent: similar national resources may translate into markedly
different outcomes across events, and the same country can exhibit strong
advantages in some sports but limited competitiveness in others. These
characteristics make Olympic forecasting a heterogeneous and nonlinear
prediction problem rather than a simple mapping from macroeconomic resources
to medal totals~\cite{bernard2004,nevill2012}.

Econometric and forecasting studies have related Olympic success to GDP,
population, host effects, public spending, and prior
performance~\cite{bernard2004,nevill2012,forrest2010,scelles2020}.
More recent work has introduced machine-learning approaches to capture
nonlinear relationships and complex interactions in Olympic
data~\cite{schlembach2022}. However, predictive flexibility alone does not
guarantee reliable or informative analysis. First, Olympic forecasting is
intrinsically temporal, yet random or contemporaneous data splits may
overestimate generalization to future Games. Second, strong persistence in
national performance makes it difficult to distinguish genuinely informative
predictors from repeated historical success or country-specific identity
effects. Third, most existing studies operate at the aggregate national level,
providing limited insight into why a country succeeds in one sport but not
another. At the same time, increasingly complex models raise an
interpretability problem. Explainable AI methods such as
LIME~\cite{ribeiro2016} and SHAP~\cite{lundberg2017} provide principled tools
for interpreting model predictions, but their use in Olympic analytics has
rarely been integrated with temporally rigorous and sport-specific
forecasting.

\textbf{Our work.}
We formulate Olympic performance as a two-level learning problem that connects
national medal capacity with sport-specific competitive advantage. At the
NOC--Games level, \textbf{Module A} models whether a country wins a medal and,
conditional on success, how many medals it wins, using socioeconomic
conditions, Games context, delegation structure, and historical performance.
At the NOC--Games--Sport level, \textbf{Module B} extends the same national
context with sport-specific participation, athlete allocation, and competitive
history to model medal success within individual sports. To make these
predictions interpretable, we incorporate SHAP-based
explainability~\cite{lundberg2017} to quantify how national- and sport-level
factors contribute to model outputs. The framework is evaluated
chronologically so that future Games are predicted exclusively from
information available in previous editions.

Empirically, the two levels reveal different but complementary structures of
Olympic performance. National medal success is associated not only with
socioeconomic capacity, but more strongly with delegation characteristics and
accumulated competitive experience, whereas sport-level success depends
substantially on how athletes, event opportunities, and prior expertise are
distributed across individual sports. The resulting analysis therefore moves
beyond aggregate medal forecasting toward explaining how broad national
capacity is converted into sport-specific competitive advantage.

Our contributions are threefold:
(\textit{i}) a two-level formulation linking national medal capacity to
sport-specific competitive advantage;
(\textit{ii}) a temporally rigorous evaluation protocol for future-Games
prediction beyond persistence and country-identity effects; and
(\textit{iii}) an explainable AI framework that reveals how national resources
and sport-specific allocation jointly translate into Olympic success.

% =============================================================================
\section{Data and Methodology}
\label{sec:methodology}
% =============================================================================

\subsection{Data Sources}
\label{subsec:data}

We integrate historical Summer Olympic athlete--event records, official
medal-table outcomes, host and sport-program information, and annual
socioeconomic indicators from the World Bank World Development Indicators
(WDI) database~\cite{worldbank2026}. The source data span 17 Summer Olympic
Games from 1960 to 2024; 1960 serves as the left boundary for constructing
lagged historical features, and the primary modeling sample begins in 1964.
Historical NOC labels and sport names are harmonized across editions, and
official medal-table records define national medal totals, avoiding duplicate
counting of team-event medals from athlete-level records. For Games year $t$,
socioeconomic predictors use only information available before competition,
taking the latest WDI observation at or before $t-1$ within a three-year
look-back window. The primary prediction horizon is immediately before each
Games, after delegation entries and the Olympic programme are known but before
competition outcomes are observed. The data are organized at two complementary
resolutions. Module A contains 2,584 NOC--Games observations from 219 NOCs and
models whether an NOC wins at least one medal and, conditional on success, its
total medal count. Module B contains 21,068 NOC--Games--Sport observations
across 41 harmonized sport families and models whether an NOC wins at least
one medal in a given sport. Paris 2024 is retained as an out-of-time test set,
comprising 204 observations in Module A and 1,876 in Module B.

% -----------------------------------------------------------------------------
\subsection{Feature Construction}
\label{subsec:features}
% -----------------------------------------------------------------------------

Features are organized into conceptually distinct blocks
(Table~\ref{tab:features}). Module A uses 14 primary predictors covering
socioeconomic context, Games conditions, delegation capacity, and historical
national performance. Delegation size is transformed using $\log(1+x)$ to
reduce right skew. Delegation breadth is measured by Shannon entropy,

\begin{equation}
H_{c,t}
=
-\sum_{s\in\mathcal{S}_{c,t}}
p_{c,t,s}\log p_{c,t,s},
\label{eq:entropy}
\end{equation}

where $p_{c,t,s}$ is the share of athletes from NOC $c$ at Games $t$
allocated to sport $s$. Historical predictors are strictly lagged:
MED\_L1 denotes the previous-Games medal count, while PG and PMR summarize
prior participation and medal-winning frequency. Pairwise Spearman
correlations show expected dependence among delegation and historical
variables; the largest absolute correlation is 0.87 between MED\_L1 and PMR.
Because no pair exceeds 0.90 and the variables represent distinct analytical
constructs, they are retained rather than removed mechanically.

\begin{table}[t]
\centering
\caption{Predictor blocks used in the two analytical modules. Compact
abbreviations are used in subsequent figures.}
\label{tab:features}
\scriptsize
\setlength{\tabcolsep}{3.0pt}

\begin{tabular}{
>{\raggedright\arraybackslash}p{1.8cm}
>{\raggedright\arraybackslash}p{2.7cm}
cc
>{\raggedright\arraybackslash}p{4.9cm}
}

\toprule

Block
& Representative variables
& A
& B
& Analytical role / construction \\

\midrule

Socioeconomic context
& lnGDP, lnPOP, URB, GDPg
& $\checkmark$
& $\checkmark$
& National economic scale, population, urbanization, and annual GDP growth;
GDP and population are log-transformed. \\

Games context
& HOST, EVT
& $\checkmark$
& $\checkmark$
& Host indicator and total events available in the Games. \\

Delegation capacity
& lnDEL, FEM, RET, PMD, SENT
& $\checkmark$
& partial
& Current delegation size and composition; SENT is the entropy of athlete
allocation across sports. \\

National history
& MED\_L1, PG, PMR
& $\checkmark$
& $\checkmark$
& Previous-Games medal count, number of prior Games, and fraction of prior
Games with at least one medal. \\

Sport allocation
& SAC, SAS, SEE, SFS, SRET, SPMD
& --
& $\checkmark$
& Sport-level athlete count/share, event entries, female share,
returning-athlete share, and prior-medalist share. \\

Sport history
& SHM\_L1, SHPG, SHMR
& --
& $\checkmark$
& Previous-Games medal status and prior participation/medal rates for the same
NOC--sport pair. \\

Sport opportunity / identity
& SGE, SNC, SPT
& --
& $\checkmark$
& Events available in the sport, number of participating NOCs, and harmonized
sport identity. \\

\bottomrule

\end{tabular}
\end{table}

Module B retains a compact national-context subset and augments it with
sport-specific allocation, participation, composition, history, and
competitive-context variables. Sport identity is treated as an unordered
categorical variable and one-hot encoded within each temporal training fold.
Current-Games medal outcomes, medal colors, post-competition variables, and
NOC identity are excluded from the primary predictors. Numerical imputation
and, where required, feature scaling are fitted within each training fold,
preventing future-Games information from entering model development.

% -----------------------------------------------------------------------------
\subsection{Modeling Framework}
\label{subsec:model}
% -----------------------------------------------------------------------------

\paragraph{Two-level prediction.}

We formulate Olympic performance as two linked prediction tasks. At the
national level, Module A separates \emph{medal attainment} from
\emph{medal accumulation}, following the general logic of two-part models for
outcomes with a substantial probability mass at zero~\cite{mullahy1986}.
For NOC $c$ at Games $t$,

\begin{equation}
\begin{aligned}
p_{c,t}
&=
f_{\theta}(\mathbf{x}_{c,t})
=
P\!\left(
Y^{\mathrm{any}}_{c,t}=1
\mid \mathbf{x}_{c,t}
\right),\\
m_{c,t}
&=
g_{\psi}(\mathbf{x}_{c,t})
\approx
\mathbb{E}\!\left[
Y^{\mathrm{medal}}_{c,t}
\mid
Y^{\mathrm{any}}_{c,t}=1,
\mathbf{x}_{c,t}
\right],\\
\widehat{Y}^{\mathrm{medal}}_{c,t}
&=
\mathbb{I}(p_{c,t}\geq\tau_A)
\max(0,m_{c,t}),
\end{aligned}
\label{eq:moduleA}
\end{equation}

where $\mathbf{x}_{c,t}$ is the national predictor vector. Stage~1 predicts
medal attainment, while Stage~2 is fitted only on medal-positive training
observations to estimate conditional medal totals. The threshold $\tau_A$ is
fixed from development predictions.

Module B extends the analysis to the NOC--Games--Sport level:

\begin{equation}
q_{c,t,s}
=
h_{\omega}(\mathbf{z}_{c,t,s})
=
P\!\left(
Y^{\mathrm{sport}}_{c,t,s}=1
\mid
\mathbf{z}_{c,t,s}
\right),
\label{eq:moduleB}
\end{equation}

where $\mathbf{z}_{c,t,s}$ combines national context with sport-specific
allocation, participation, history, and competitive opportunity. Thus,
Module A estimates national medal capacity, whereas Module B characterizes
how that capacity is expressed across sports.

\paragraph{Model selection and temporal learning.}

Stage~1 and Module B compare nine classifiers: LR, RF~\cite{breiman2001},
SVM~\cite{cortes1995}, KNN, BPNN, DT, XGBoost~\cite{chen2016},
CatBoost~\cite{prokhorenkova2018}, and Gaussian NB. Stage~2 compares linear
regression, Ridge, RF, SVR, KNN, BPNN, DT, XGBoost, and CatBoost.
Models are selected solely by expanding-window development performance.
Algorithm~\ref{alg:framework} summarizes the complete pipeline.

\begin{algorithm}[t]
\caption{Temporal two-level learning and interpretation pipeline}
\label{alg:framework}
\small

\begin{algorithmic}[1]

\Require National panel $\mathcal{D}_A$, sport panel $\mathcal{D}_B$,
candidate model sets, development Games $\mathcal{T}$

\Ensure 2024 national and sport-level predictions with post-hoc explanations

\For{each validation Games $t\in\mathcal{T}$}
    \State Fit preprocessing using observations strictly before $t$
    \State Fit Module-A Stage~1 classifiers on training NOC--Games
    \State Fit Stage~2 regressors on medal-positive training observations
    \State Fit Module-B classifiers on training NOC--Games--Sport observations
    \State Evaluate classification by AP and Stage~2 regression by RMSE
\EndFor

\State Select the best model family for each task by mean temporal performance
\State Lock $\tau_A$ and the Module-B threshold using development predictions
\State Refit selected models using all observations available through 2020
\State Predict withheld Paris 2024 with
Eqs.~\eqref{eq:moduleA}--\eqref{eq:moduleB}

\State Fit performance-matched tree reference models for SHAP analysis
\State Compute global and signed feature attributions
\State Estimate nonlinear SHAP responses and zero-crossing thresholds
\State Cluster 2024 NOCs into capacity profiles and compute profile-centered
sport contrasts

\end{algorithmic}
\end{algorithm}

\paragraph{Explainable AI and nonlinear responses.}

SHAP characterizes additive feature contributions to model outputs
~\cite{lundberg2017,lai2024}. For an observation $\mathbf{x}$,

\begin{equation}
f(\mathbf{x})
=
\phi_0+\sum_{j=1}^{M}\phi_j(\mathbf{x}),
\label{eq:shap}
\end{equation}

where $\phi_0$ is the expected output and $\phi_j$ is the contribution of
feature $j$. Mean absolute SHAP summarizes global relevance, while signed
values preserve contribution direction.

Because the selected predictors are not necessarily tree-based,
TreeSHAP~\cite{lundberg2020} is applied to independently fitted tree models
with closely matched temporal performance. LOWESS~\cite{cleveland1979}
summarizes nonlinear SHAP responses; a zero crossing is treated as a
\emph{model-implied effect threshold}, indicating a sign change in average
contribution relative to the model baseline. These thresholds describe
predictive, not causal, effects. SHAP magnitudes are interpreted within each
module because Modules A and B use different reference models and output
spaces.

\paragraph{Country profiles and relative sport advantage.}

To examine cross-sport heterogeneity, the 2024 NOCs are grouped into four
descriptive profiles using $k$-means on standardized national resource,
delegation, and historical characteristics~\cite{macqueen1967}. Profiles are
used only for post-hoc interpretation. For profile $p$ and sport $s$, the
country-context SHAP contribution is centered by the profile's
participation-weighted cross-sport baseline:

\begin{equation}
\mathrm{CA}_{p,s}
=
\overline{\Phi}^{\mathrm{country}}_{p,s}
-
\frac{
\sum_r n_{p,r}\,
\overline{\Phi}^{\mathrm{country}}_{p,r}
}{
\sum_r n_{p,r}
},
\label{eq:ca}
\end{equation}

where $\overline{\Phi}^{\mathrm{country}}_{p,s}$ is the mean summed
country-context contribution and $n_{p,s}$ is the corresponding cell size.
Positive values indicate a more favorable contribution than the same
profile's cross-sport baseline.

% -----------------------------------------------------------------------------
\subsection{Evaluation Protocol}
\label{subsec:evaluation}
% -----------------------------------------------------------------------------

We use expanding-window validation to reproduce future-Games forecasting
using only historically available information~\cite{tashman2000}. The
development validation editions are 2008, 2012, 2016, and 2020, while Paris
2024 is withheld from model selection, preprocessing, and threshold tuning.

Average precision (AP) is the primary classification metric under class
imbalance~\cite{davis2006,saito2015}, with ROC-AUC, F1, MCC, accuracy,
balanced accuracy, and Brier score reported as complementary diagnostics.
Stage~2 is evaluated by RMSE and MAE, with $R^2$ reported secondarily.
Classification thresholds maximize MCC on pooled development predictions and
remain fixed for 2024. Robustness is assessed against previous-Games
persistence, feature-block ablation, and an NOC-identity sensitivity test;
NOC identity is excluded from the primary predictive and explainable models.

% =============================================================================
\section{Results and Discussion}
\label{sec:results}
% =============================================================================

\subsection{Predictive Performance}
\label{subsec:performance}

\subsubsection{Temporal Model Selection.}

Table~\ref{tab:modelcomparison} reports model performance over the four
expanding-window validation folds. The preferred learner varies across the
three prediction tasks. For national medal attainment, LR achieves the highest
mean AP ($0.9569\pm0.0083$), narrowly outperforming RF, CatBoost, and XGBoost.
For medal accumulation among medal-winning NOCs, XGBoost obtains the lowest
development RMSE ($4.898\pm0.751$ medals). At the finer NOC--sport level,
BPNN achieves the highest mean AP ($0.7825\pm0.0297$).

The contrast is informative. Under the same temporal protocol, national
medal attainment is best captured by a comparatively simple decision surface,
whereas nonlinear learners are selected for conditional medal counts and
sport-level success. Model complexity is therefore task-dependent rather than
uniformly beneficial.

\begin{table}[t]
\centering
\caption{Temporal development performance over four expanding-window folds.
Higher AP is better; lower RMSE is better. Bold values indicate the selected
model for each task.}
\label{tab:modelcomparison}
\scriptsize
\setlength{\tabcolsep}{4.0pt}

\begin{tabular}{lcc|lc}

\toprule

\multicolumn{3}{c|}{Classification: AP (mean$\pm$SD)}
& \multicolumn{2}{c}{Module A Stage~2} \\

Model
& Module A
& Module B
& Regressor
& RMSE (mean$\pm$SD) \\

\midrule

LR
& \textbf{0.957$\pm$0.008}
& 0.755$\pm$0.032
& XGB
& \textbf{4.898$\pm$0.751} \\

RF
& 0.955$\pm$0.006
& 0.776$\pm$0.020
& CB
& 5.438$\pm$0.683 \\

SVM
& 0.949$\pm$0.008
& 0.755$\pm$0.032
& RF
& 5.550$\pm$1.126 \\

KNN
& 0.941$\pm$0.009
& 0.743$\pm$0.012
& BPNN
& 5.702$\pm$1.705 \\

BPNN
& 0.951$\pm$0.010
& \textbf{0.783$\pm$0.030}
& Ridge
& 5.779$\pm$1.508 \\

DT
& 0.879$\pm$0.039
& 0.667$\pm$0.033
& Linear
& 5.780$\pm$1.506 \\

XGB
& 0.952$\pm$0.007
& 0.772$\pm$0.019
& DT
& 6.032$\pm$1.278 \\

CB
& 0.954$\pm$0.007
& 0.778$\pm$0.022
& KNN
& 6.397$\pm$1.779 \\

NB
& 0.939$\pm$0.021
& 0.611$\pm$0.027
& SVR
& 8.741$\pm$2.061 \\

\bottomrule

\end{tabular}
\end{table}

\subsubsection{External Validation and Error Analysis.}

\begin{figure}[t]
\centering

\safeincludegraphics[
    width=0.95\linewidth,
    keepaspectratio
]{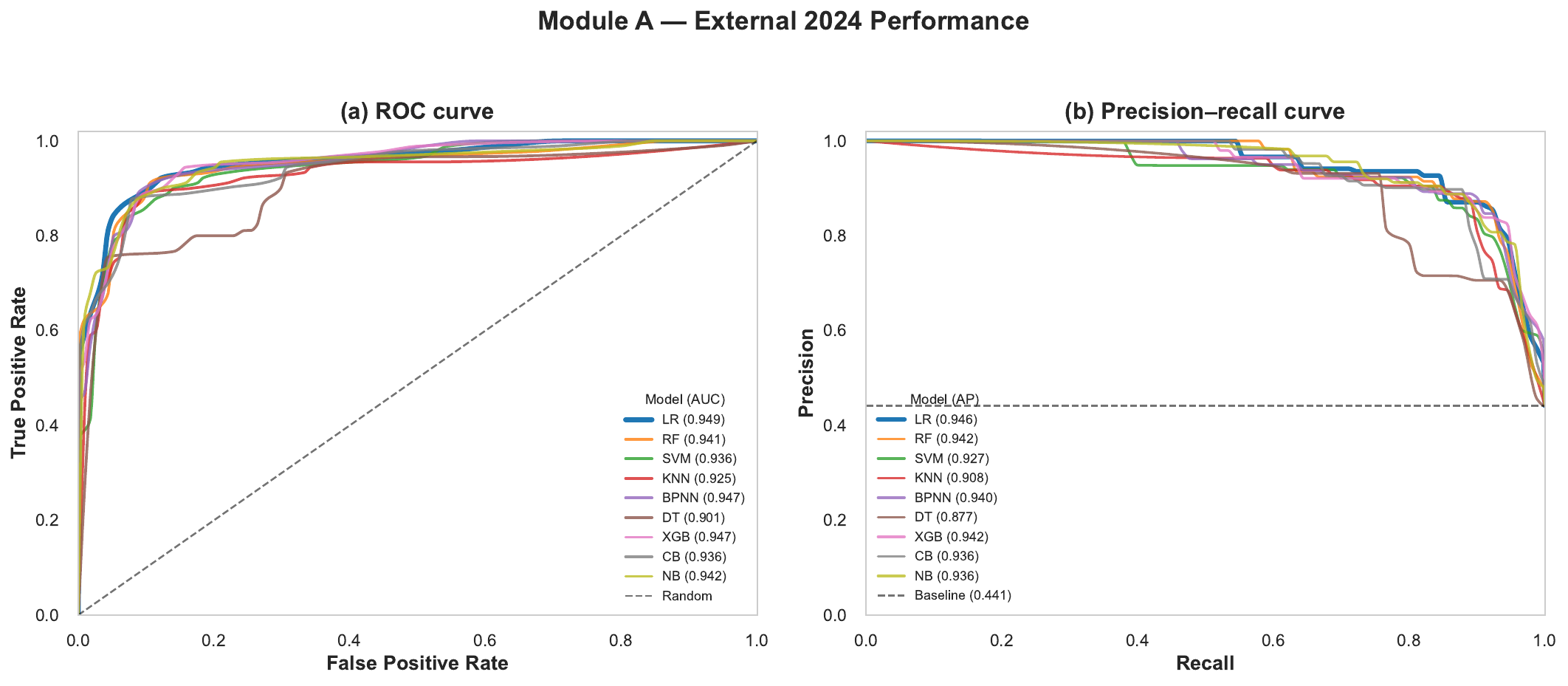}

\vspace{1mm}

\safeincludegraphics[
    width=0.95\linewidth,
    keepaspectratio
]{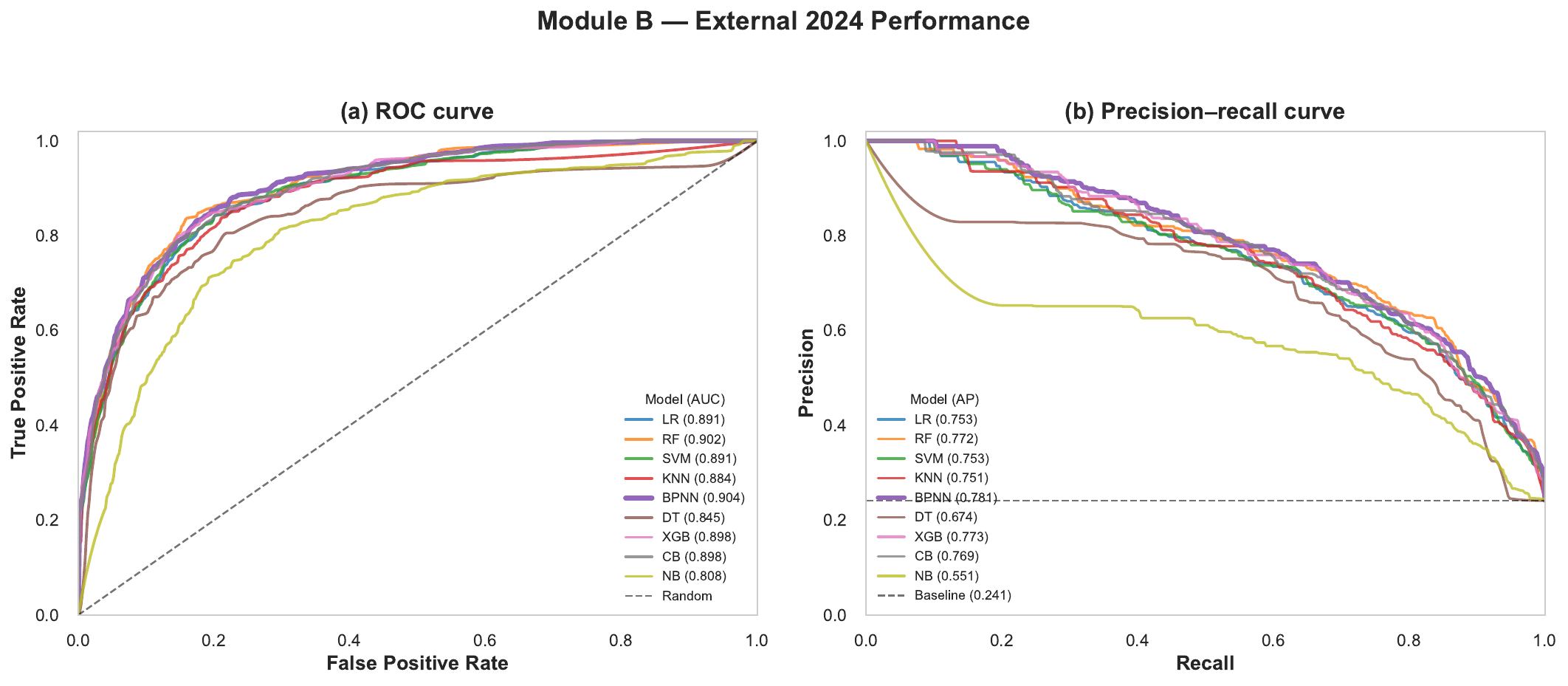}

\caption{External Paris-2024 discrimination for Module A (top) and
Module B (bottom). All AP and ROC-AUC values are computed from the original
prediction scores.}
\label{fig:external}

\end{figure}

The model ranking transfers well to the untouched Paris 2024 Games
(Table~\ref{tab:external2024}). Module A LR achieves AP=0.9464 and
ROC-AUC=0.9487 on 204 NOCs, compared with a development AP of 0.9569.
Module B shows similarly limited temporal degradation: BPNN reaches
AP=0.7809 and ROC-AUC=0.9036 on 1,876 NOC--sport observations, almost
identical to its development AP of 0.7825. The close development--test
agreement supports temporal stability of the selected models into the 2024
cycle (Fig.~\ref{fig:external}).

For medal counts, XGBoost attains RMSE=3.894 and MAE=2.487 among the
90 true medal-winning NOCs. When combined with the locked Stage~1 decision
rule, the complete two-stage system reaches an end-to-end RMSE of 2.659,
MAE of 1.248, and $R^2=0.9653$ across all NOCs. The classification gate
identifies 76 of the 90 medal-winning NOCs, with eight false positives and
14 false negatives.

\begin{table}[t]
\centering
\caption{Performance on the withheld Paris 2024 Games.}
\label{tab:external2024}
\scriptsize
\setlength{\tabcolsep}{4.0pt}

\begin{tabular}{llrrrr}

\toprule

Task
& Model / output
& AP
& ROC-AUC
& F1
& MCC \\

\midrule

A: medal attainment
& LR
& 0.946
& 0.949
& 0.874
& 0.781 \\

B: sport medal status
& BPNN
& 0.781
& 0.904
& 0.694
& 0.601 \\

\midrule

\multicolumn{2}{l}{Count evaluation}
& RMSE
& MAE
& \multicolumn{2}{c}{$R^2$} \\

\midrule

A: true winners
& XGBoost
& 3.894
& 2.487
& \multicolumn{2}{c}{0.961} \\

A: end-to-end
& Continuous
& 2.659
& 1.248
& \multicolumn{2}{c}{0.965} \\

\bottomrule

\end{tabular}
\end{table}

The actual--predicted comparison further exposes where the remaining errors
occur. The model predicts 110.1 medals for the USA
compared with 126 observed, 80.0 versus 91 for China, 51.1 versus 64 for
France, and 17.8 versus 32 for South Korea. The continuous predictions sum
to 996.7 medals compared with 1,038 observed, indicating mild aggregate
underprediction that is concentrated mainly among high-medal NOCs and
breakthrough medal winners. Thus, high overall $R^2$ does not imply uniform
accuracy across the outcome distribution.

% =============================================================================
\subsection{National Medal Capacity}
\label{subsec:national}
% =============================================================================

\subsubsection{Medal Attainment and Accumulation.}

Feature-block ablations reveal that entering the medal-winning set and
accumulating medals are related but not identical prediction problems
(Table~\ref{tab:ablation}). For Stage~1, removing delegation information
reduces AP from 0.9569 to 0.9413, a larger deterioration than removing
national history (0.9495). By contrast, in Stage~2, removing historical
performance raises RMSE from 4.898 to 6.739, compared with 5.656 after
removing delegation variables.

This difference suggests a meaningful separation between \emph{medal
attainment} and \emph{medal accumulation}. Current delegation composition
provides a strong signal of whether an NOC possesses sufficient competitive
breadth to cross the medal-entry boundary, whereas accumulated Olympic
experience becomes increasingly important once medal-winning capacity has
already been established.

A macro/Games-only model remains informative but performs substantially worse:
AP falls to 0.8382 for attainment and RMSE rises to 14.211 for medal
accumulation. This should not be interpreted as evidence that socioeconomic
conditions are unimportant. Rather, macroeconomic conditions provide a broad
background capacity signal whose predictive value is sharpened by delegation
formation and accumulated sporting experience.

\begin{table}[t]
\centering
\caption{Feature-block ablations over the temporal development folds.
Higher AP and lower RMSE indicate better performance.}
\label{tab:ablation}
\scriptsize

\setlength{\tabcolsep}{2.6pt}
\renewcommand{\arraystretch}{0.92}

\begin{tabular}{
@{}lr
@{\hspace{0.8em}}
lr
@{\hspace{0.8em}}
lr@{}
}

\toprule

\multicolumn{2}{c}{A Stage~1 (AP)}
&
\multicolumn{2}{c}{A Stage~2 (RMSE)}
&
\multicolumn{2}{c}{Module B (AP)}
\\

\cmidrule(r){1-2}
\cmidrule(lr){3-4}
\cmidrule(l){5-6}

Specification & Score
& Specification & Score
& Specification & Score
\\

\midrule

Full
& 0.9569
& Full
& 4.898
& Full
& 0.7825 \\

No delegation
& 0.9413
& No delegation
& 5.656
& No sport allocation
& 0.7195 \\

No history
& 0.9495
& No history
& 6.739
& No sport history
& 0.7730 \\

Macro/Games only
& 0.8382
& Macro/Games only
& 14.211
& Country context only
& 0.4851 \\

&
&
&
& Sport-specific only
& 0.7798 \\

\bottomrule

\end{tabular}
\end{table}

\subsubsection{Feature Attribution and Nonlinear Responses.}

The SHAP analysis sharpens this interpretation from feature blocks to
individual effects. Delegation size (lnDEL) produces the largest mean absolute
SHAP value (0.137), followed by previous-Games medals (MED\_L1, 0.078),
historical medal-Games rate (PMR, 0.076), prior-medalist share (PMD, 0.054),
and sport entropy (SENT, 0.054). GDP remains the leading purely economic
variable (0.030). Hence, economic scale is not the strongest immediate
predictor, but it remains a systematic component of the national capacity
structure.

\begin{figure}[t]
\centering

\safeincludegraphics[
    width=0.98\linewidth,
    keepaspectratio
]{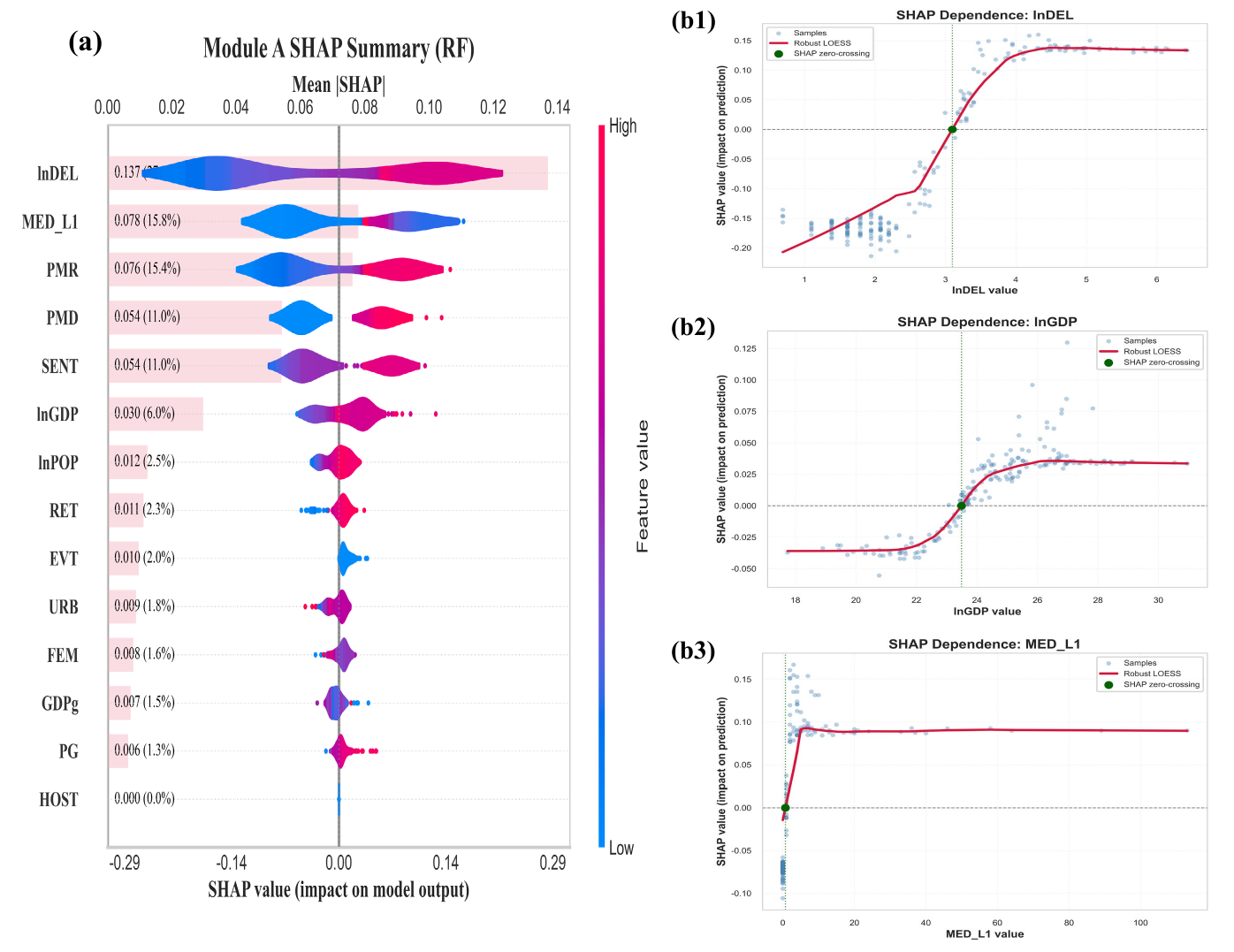}

\caption{Feature attribution and nonlinear responses in Module A.
Panel (a) shows the SHAP summary and mean absolute importance.
Panels (b1)--(b3) show the nonlinear responses of delegation size,
GDP, and previous-Games medals, respectively.}
\label{fig:shapa}

\end{figure}

Importantly, global importance alone obscures the direction and shape of these
effects. In Fig.~\ref{fig:shapa}, the sign of a SHAP value indicates whether
a feature raises or suppresses the model output relative to its baseline.
We therefore interpret the zero crossing of each smoothed response as a
\emph{model-implied effect threshold}. For delegation size, the threshold
occurs at $\mathrm{lnDEL}=3.10$, equivalent to approximately 21 athletes.
Delegations below this level exert a predominantly negative predictive effect
on medal probability; once the threshold is crossed, the contribution becomes
positive and rises rapidly before flattening for larger delegations.
The shape therefore combines a threshold effect with diminishing marginal
predictive gains.

GDP exhibits a similarly informative but smoother nonlinear structure.
Its SHAP response crosses zero at $\mathrm{lnGDP}=23.49$, corresponding to
approximately US\$15.9 billion. Below this level, GDP contributes negatively
to the national medal prediction relative to the model baseline; above it,
the contribution becomes positive. The effect grows most rapidly across the
middle of the GDP distribution and subsequently approaches a plateau at high
income scales. Thus, the model does not imply that economic resources are
irrelevant or that their contribution grows linearly without bound. Instead,
GDP displays both an \emph{effect threshold} and a \emph{saturation pattern}:
sufficient economic scale appears to provide an enabling condition for
competitive success, after which additional economic size yields progressively
smaller changes in predicted medal probability.

Previous-Games medals (MED\_L1) display a distinct step-like response.
Its SHAP contribution rises sharply from the low-medal range and quickly
reaches a relatively stable positive plateau. This indicates that recent medal
experience captures a strong persistence effect: once a country has established
medal-winning capability, additional previous-Games medals generate only
limited marginal increases in predicted medal probability.

Taken together, these results distinguish resource availability from resource
conversion and accumulated competitive experience. GDP defines part of the
socioeconomic foundation, whereas delegation size, prior medal experience,
and sport breadth more directly describe how that foundation is converted into
Olympic competitiveness. This extends the conventional resource-based
explanation of Olympic success by showing that the predictive effects are
strongly nonlinear and operate through different stages of competitive
capacity~\cite{bernard2004,scelles2020}.

% =============================================================================
\subsection{Sport-Specific Advantage}
\label{subsec:sport}
% =============================================================================

\subsubsection{Sport-Level Predictive Drivers.}

Moving from the national to the NOC--sport level changes both the relative
importance and the functional form of the predictive factors. The
sport-specific-only specification reaches AP=0.7798, close to the full model
(0.7825), whereas country context alone falls to 0.4851. Removing sport
allocation produces the largest deterioration (AP=0.7195). These results
show that national capacity alone cannot determine \emph{where} medals are won;
the conversion of national resources into medals is primarily resolved at
the sport level.

The conceptual SHAP ranking reinforces this result. Sport athlete count (SAC)
has the largest mean absolute contribution (0.853), followed by sport event
entries (SEE, 0.482), same-sport historical medal rate (SHMR, 0.401), overall
delegation size (0.291), national historical medal rate (0.254), and
previous-Games same-sport medal status (0.239). Sport identity itself also
carries substantial information (0.227). Hence, sport allocation, sport
history, and the type of sport jointly determine how broad national capacity
is expressed at the competitive level. These SHAP magnitudes are interpreted
within Module B and are not numerically comparable with Module A.

\begin{figure}[t]
\centering

\safeincludegraphics[
    width=0.98\linewidth,
    height=8.0cm,
    keepaspectratio
]{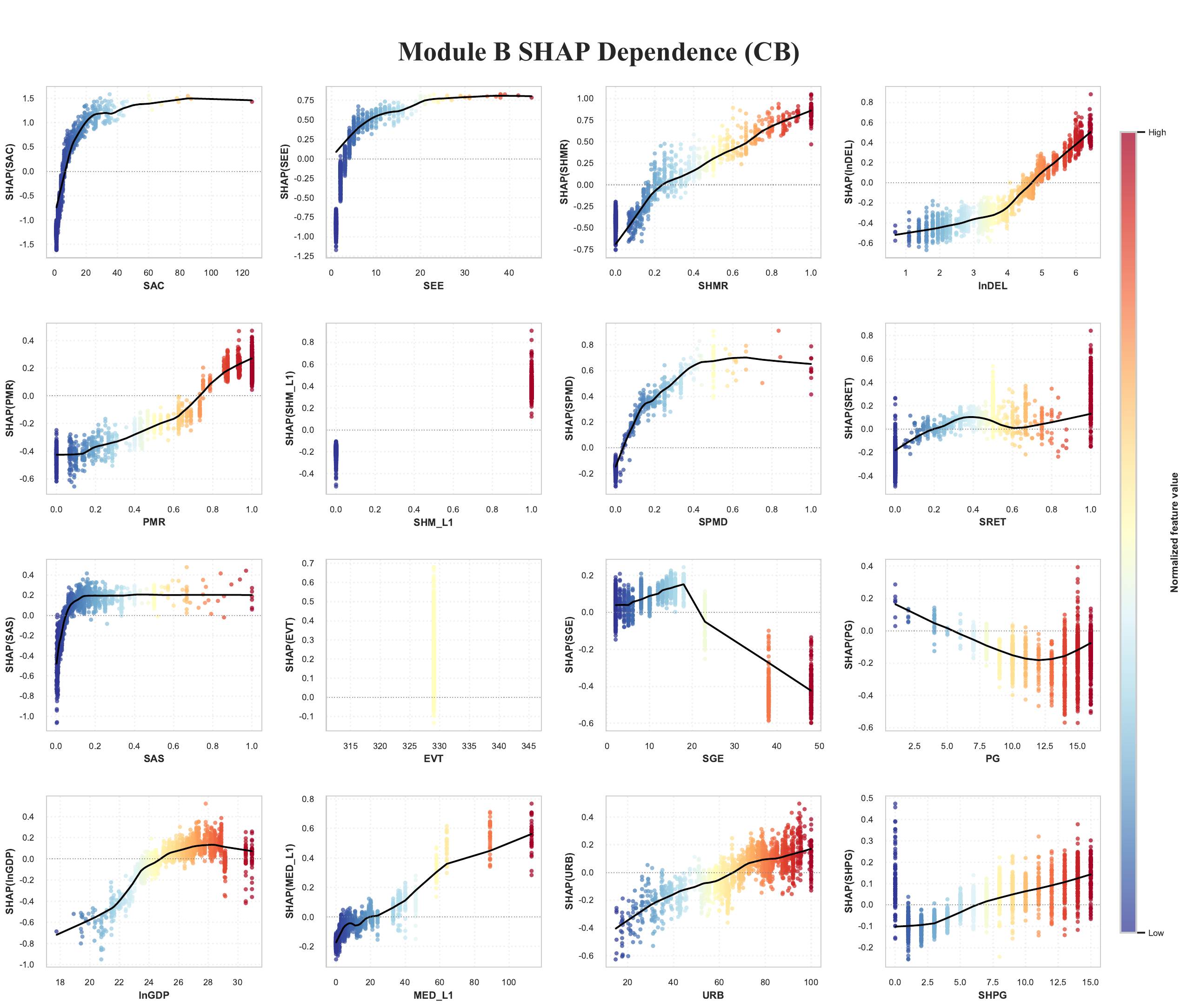}

\caption{Nonlinear SHAP dependence patterns in Module B.
Positive and negative values indicate upward and downward effects on the
model output relative to its baseline. Several variables exhibit threshold,
saturation, or non-monotonic effects, demonstrating that sport-specific
competitive advantage cannot be reduced to linear resource scaling.
EVT is constant across the Paris 2024 observations (329 events), whereas
SHM\_L1 is binary; LOWESS curves are therefore omitted for both variables.}
\label{fig:sportdependence}

\end{figure}

More importantly, the dependence responses in
Fig.~\ref{fig:sportdependence} reveal that these effects are not uniformly
linear. SAC crosses the zero-SHAP line at approximately 6.57 athletes:
sport delegations below roughly seven athletes tend to exert a negative
effect on the model output, whereas larger allocations contribute positively.
The curve subsequently flattens, again revealing diminishing predictive
returns to scale. SHMR crosses from negative to positive at approximately
0.239, indicating that a sufficiently established same-sport medal history
changes the direction of its contribution. At the national level within
Module B, lnDEL crosses zero at 4.70, equivalent to an overall delegation of
approximately 109 athletes, after which national delegation capacity provides
a positive contribution to sport-level success.

Other features display equally important nonlinearities. Event entries rise
rapidly from negative to positive contributions before approaching saturation,
while prior-medalist share increases strongly at low-to-moderate levels and
then levels off. By contrast, several responses are non-monotonic: the effects
of returning-athlete share and sport-event availability can strengthen over
one part of the feature range and weaken over another. GDP also retains a
nonlinear positive region at the sport level, but with smaller magnitude than
direct sport-allocation variables. These patterns explain why simple linear
resource-performance relationships are insufficient: the sign and magnitude
of a factor's contribution depend on where a country or sport lies within the
observed feature range.

\subsubsection{Country Profiles and Comparative Advantage.}

The feature-level results describe average behavior across NOCs, but the
same national resources need not generate the same effect in every sport.
We therefore examine heterogeneity across national capacity profiles and sport
contexts. The 2024 NOCs form four descriptive profiles containing 51, 77, 65,
and 11 countries. The profiles span a broad capacity gradient, with Profile~1
at the lower-capacity end and Profile~4 at the higher-capacity end, while
Profiles~2--3 represent intermediate but distinct configurations. Rather than
comparing raw SHAP levels---which would largely reproduce overall differences
in national capacity---we center country-context SHAP within each profile.

\begin{figure}[t]
\centering

\safeincludegraphics[
    width=0.98\linewidth,
    height=5.8cm,
    keepaspectratio
]{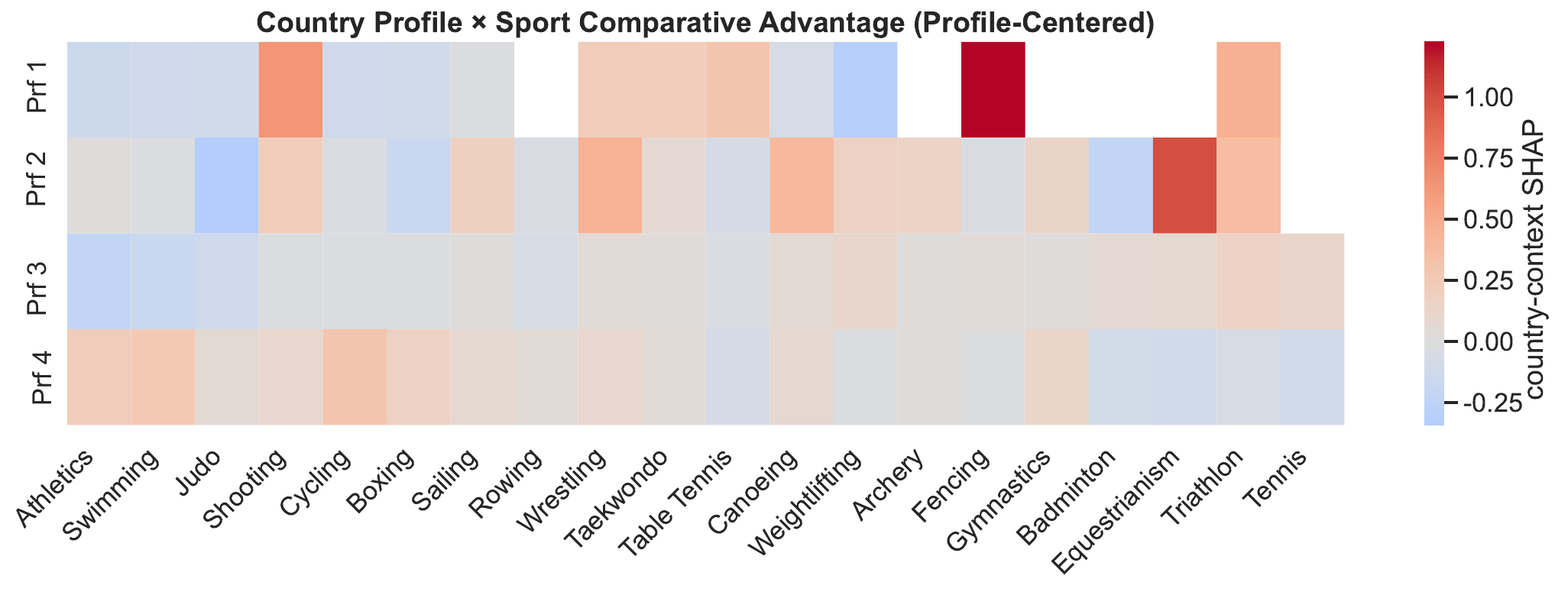}

\caption{Country-profile $\times$ sport comparative advantage in Paris 2024.
Values are profile-centered country-context SHAP contributions. Positive
values indicate sports receiving a more favorable national-context effect
than the profile's own cross-sport baseline, while negative values indicate
a less favorable relative effect.}
\label{fig:profilesport}

\end{figure}

The resulting heatmap in Fig.~\ref{fig:profilesport} reveals clear
profile--sport heterogeneity. Among cells with at least ten
observations, Profile~4 shows its strongest positive relative effects in
cycling (0.299), swimming (0.269), athletics (0.228), and boxing (0.172).
Profile~3 instead displays positive comparative effects in triathlon
(0.165), tennis (0.123), weightlifting (0.109), and equestrianism (0.082).
Profile~2 is relatively advantaged in shooting (0.215) and taekwondo (0.075),
whereas Profile~1 shows no positive centered effect among sports meeting the
same sample-size criterion.

These differences are important because the centered score controls for each
profile's own overall capacity level. A strong national resource base therefore
does not translate into a uniform positive effect across the Olympic program.
Instead, the contribution of national context is conditional on the sport in
which that capacity is deployed. In this sense, relative advantage reflects
the combination of general national capacity and sport-specific competitive
structures rather than economic strength alone.

This profile--sport heterogeneity also provides the applied interpretation of
Module B. Countries with similar overall capacity may occupy different
sport-specific opportunity spaces, implying that benchmarking based solely on
GDP, population, or total medals obscures meaningful specialization. The
centered SHAP framework makes these differences explicit and provides a
data-driven basis for comparing where a national competitive profile is
relatively more or less favorable.

% =============================================================================
\subsection{Robustness Analysis}
\label{subsec:robustness}
% =============================================================================

The main findings remain stable across three complementary robustness checks.
First, the proposed framework consistently outperforms previous-Games
persistence: national medal-status AP increases from 0.7662 to 0.9464,
end-to-end medal-count RMSE decreases from 3.256 to 2.659 ($18.3\%$), and
sport-level AP rises from 0.5172 to 0.7809. Second, adding explicit NOC
identity changes the 2024 AP only marginally (0.9464 to 0.9475), indicating
that the national ranking is driven primarily by measured structural,
delegation, and historical features rather than country labels. Third, the
tree-based reference models used for SHAP interpretation remain closely matched
to the primary predictive models, with development AP gaps of only 0.0022 for
Module A and 0.0041 for Module B. These results indicate that the predictive
gains and corresponding interpretation are not artifacts of historical
carry-over, country-identity memorization, or substantially weaker explanation
models.

% =============================================================================
\section{Conclusion}
\label{sec:conclusion}
% =============================================================================

We propose a temporally validated and explainable two-level framework linking
national Olympic medal capacity with sport-specific competitive advantage. On
the withheld Paris 2024 test set, the national classifier achieves AP=0.9464 and
the complete two-stage model reaches an end-to-end RMSE of 2.659 medals,
reducing RMSE by 18.3\% relative to previous-Games persistence; at the sport
level, AP rises from 0.5172 to 0.7809. The two analytical levels reveal a
consistent conversion structure: delegation characteristics are particularly
informative for medal attainment, accumulated competitive history for medal
accumulation, and sport-specific allocation and history for where medals are
won. SHAP dependence further identifies nonlinear capacity transitions, with
GDP and delegation size changing from negative to positive contributions at
approximately US\$15.9 billion and 21 athletes, respectively, and sport athlete
count at roughly seven athletes. Country profiling shows that similar broad
capacity does not translate uniformly across sports, revealing distinct
profile-specific relative advantages. Overall, Olympic competitiveness depends
not only on national resources, but on how those resources and accumulated
expertise are converted into delegation capacity and sport-specific
specialization. This provides an interpretable basis for benchmarking and
resource-allocation analysis in elite sport.

The analysis is observational, so SHAP effects and model-implied thresholds
should not be interpreted as causal intervention effects. Future work will
incorporate richer institutional and athlete-level information and test the
stability of these thresholds and profile--sport patterns across subsequent
Olympic cycles.

% =============================================================================
\section*{Acknowledgements}
% =============================================================================

This work was supported by the Key Laboratory of Social Computing and
Intelligent Governance in Western Sichuan Plateau at Sichuan Minzu College
(Grant No. CXGY202608), through the project ``Research on the Measurement and
Enhancement Strategies of the Resilience of the Agro-pastoral Industry Chain
and Supply Chain in the Western Sichuan Plateau under Extreme Climate
Disturbances'', and the Scientific Research Start-up Project for Introduced
Talents of Chengdu University of Information Technology, ``Research on the
Benefit-Sharing and Compensation Mechanism for Coordinated Development of the
Chengdu--Chongqing Economic Circle'' (Grant No. KYTZ202298).

% =============================================================================
% References
% =============================================================================

\end{document}